%% file: root.tex
\documentclass[letterpaper, 10 pt, journal, twoside]{IEEEtran}  

\IEEEoverridecommandlockouts                              

\usepackage{amsmath}
\usepackage{mathtools}

\usepackage{pgfplots}
\usepackage{pdfpages}
\usepackage{gensymb}
\usepackage{tikz}
\usepackage{pgf}
\usepackage{mathrsfs}
\usetikzlibrary{shapes,arrows}
\usepackage{multirow}
\usepackage{multicol}
\usepackage{url}
\usepackage{multicol}
\usepackage{algorithm}
\usepackage{algpseudocode}
\usepackage{eqparbox,xparse}
\usepackage{soul}

\usepackage{bm}
\usepackage{acro}
\input{include/acronyms.tex}

\input{include/newcommands}
\usepackage{comment}
\usepackage{siunitx}
\newcommand{\vect}[1]{\boldsymbol{\mathbf{#1}}}
\usepackage{subcaption}
\usepackage{cite}
\usepackage{booktabs}

\usepackage[normalem]{ulem}
\usepackage{xcolor}
\usepackage{cancel} 

\newboolean{show_revisions}
\setboolean{show_revisions}{False}

\newcommand{\revision}[2][]{%
  \ifthenelse{\boolean{show_revisions}}%
    {{\color{red}\sout{#1}}{\color{blue}#2}}%
    {#2}%
}

\newcommand{\mathrevision}[2][]{%
  \ifthenelse{\boolean{show_revisions}}%
    {{\color{red}\cancel{#1}}\,{\color{blue}#2}}%
    {#2}%
}

\title{\LARGE \bf
\revision[]{Towards} Robust Optimization-Based \\Autonomous Dynamic Soaring with a Fixed-Wing UAV
}

\author{Marvin Harms$^{1,2}$, Jaeyoung Lim$^{2}$, David Rohr$^{2}$, Friedrich Rockenbauer$^{2}$, Nicholas Lawrance$^{2,3}$, Roland Siegwart$^{2}$

\thanks{Manuscript received: November, 30, 2025; Revised February, 17, 2026; Accepted March, 20, 2026. This paper was recommended for publication by
Editor Soon-Jo Chung upon evaluation of the Associate Editor and Reviewers
comments. This work was supported by ETH Research Grant AvalMapper ETH-10 20-1.}%
\thanks{$^{1}$ Autonomous Robots Lab, NTNU, Norway {\tt \footnotesize marvin.c.harms@ntnu.no}}%
\thanks{$^{2}$ Autonomous Systems Lab, ETH Z\"urich, Z\"urich 8092, Switzerland {\tt \footnotesize \{jalim, drohr, rsiegwart\}@ethz.ch}}%
\thanks{$^{3}$ CSIRO Robotics, Data61, QLD 4069, Australia, { \tt\footnotesize nicholas.lawrance@csiro.au}}%
}

\usepackage{eso-pic}

\newcommand{\IEEEcopyrightnotice}{%
  \AddToShipoutPictureBG*{
    \AtPageLowerLeft{%
      \hspace{0.65in} 
      \raisebox{0.3in}{
        \begin{minipage}{\textwidth}
          \footnotesize
          © 2026 IEEE. Personal use of this material is permitted. Permission from IEEE must be obtained for all other uses, in any current or future media, including reprinting/republishing this material for advertising or promotional purposes, creating new collective works, for resale or redistribution to servers or lists, or reuse of any copyrighted component of this work in other works.
        \end{minipage}
      }%
    }%
  }%
}

\begin{document}

\markboth{IEEE Robotics and Automation Letters. Accepted March, 2026}
{Harms \MakeLowercase{\textit{et al.}}: Towards Robust Optimization-Based Autonomous Dynamic Soaring with
a Fixed-Wing UAV}

\maketitle
\IEEEcopyrightnotice 

\pagestyle{headings}

\begin{abstract}
Dynamic soaring is a flying technique to exploit the energy available in wind shear layers, enabling potentially unlimited flight without the need for internal energy sources. We propose a framework for autonomous dynamic soaring with a fixed-wing \ac{UAV}. The framework makes use of an explicit representation of the wind field and a classical approach for guidance and control of the \ac{UAV}. Robustness to wind field estimation error is achieved by constructing point-wise robust reference paths for dynamic soaring and the development of a robust path following controller for the fixed-wing \ac{UAV}. \revision[The framework is evaluated in dynamic soaring scenarios in simulation and real flight tests]{Wind estimation and path tracking performance are validated with real flight tests to demonstrate robust path-following in real wind conditions}. In simulation, we demonstrate robust dynamic soaring flight subject to varied wind conditions, estimation errors and disturbances. \revision[Critical components of the framework, including energy predictions and path-following robustness, are further validated in real flights to assure small sim-to-real gap.]{} Together, our results strongly indicate the ability of the proposed framework to achieve autonomous dynamic soaring flight in wind shear.
\end{abstract}

\section{INTRODUCTION}
Some bird species, such as the \emph{Wandering albatross} can traverse over \SI{5000}{\kilo\metre} in five days relying on wind energy alone~\cite{sachs2019maximum}. 
Heart rate monitoring and direct observation suggest that they use dynamic soaring to extract energy from the wind shear that occur over the ocean~\cite{pennycuick_1982, weimerskirch_2000}.
Dynamic soaring and static soaring are the two primary methods to extract energy from the atmosphere. Static soaring utilizes rising air (often thermal updrafts) to gain energy with altitude, dynamic soaring extracts energy by traversing wind shear layers, i.e. vertical variations of the wind speed.

Extracting energy from the atmosphere would, analogously, provide a virtually unlimited range for \iac{UAV}, enabling extended-range missions and long-endurance persistent monitoring. Specifically, autonomous dynamic soaring with \acp{UAV} has been gathering interest, as an alternative to thermal soaring as wind shears may be more widely available in the atmosphere than localized updrafts.
Unlike thermal soaring, which has been demonstrated in multiple works~\cite{allen_guidance_2008, depenbusch_autosoar_2018}, however, full autonomous dynamic soaring has not yet been demonstrated on a real \ac{UAV}. Although there is a broad body of research on the modeling and optimization of dynamic soaring trajectories~\cite{zhao2004optimal}, only few lines of research consider actual demonstration for dynamic soaring in realistic wind fields. 
Works such as~\cite{bird2014closing, bronz_flight_2021} have demonstrated deployments of wind estimation and following an optimized path. However, to the best of our knowledge, energy-positive, sustained autonomous dynamic soaring has yet to be demonstrated.

\begin{figure}[t]
    \centering
    \includegraphics[width=\linewidth]{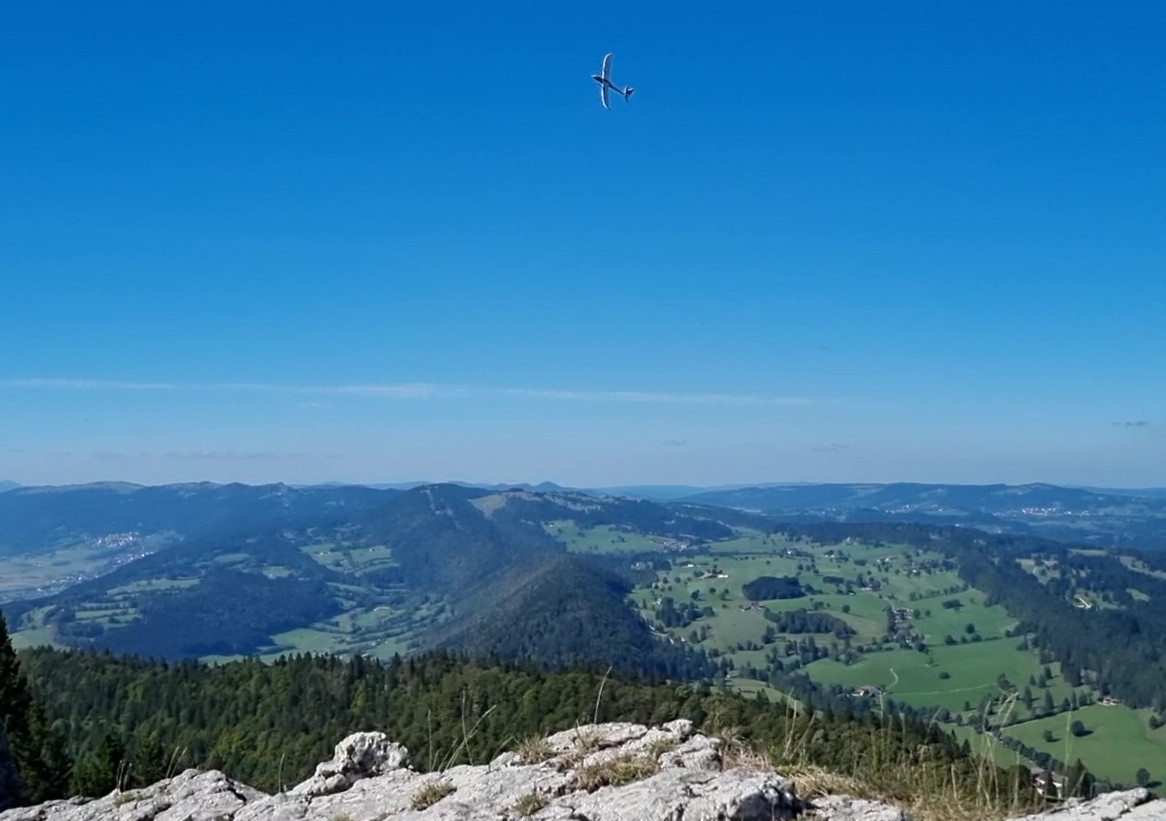}
    \caption{Impression of one of the flight tests to validate tracking performance in strong wind, conducted at Chasseral, Switzerland.}
    \label{fig:overview}
\end{figure}

In this work, we present a complete autonomous dynamic soaring framework and validate critical components in flight tests. We propose an estimation and control architecture that is robust against perturbations in the wind field (actual or estimated), as we believe this is key to achieving autonomous flight in shear. Our framework consists of three main components. First, we estimate the wind shear magnitude and extent using a filter-based parametric estimator. Second, the system generates an optimized soaring trajectory that maximizes energy gain and is robust to wind variations. Finally, we use an \ac{INDI} based path following controller to dynamically track the optimized dynamic soaring trajectory.
While we are not able to provide a full demonstration of dynamic soaring within this work, we validate the key components of our framework in a realistic dynamic soaring scenario, a step forward toward a real-world demonstration of dynamic soaring.

\IEEEpubidadjcol 

The main contributions of this research can be summarized as:
\begin{itemize}
    \item A path optimization method for robust dynamic soaring
    \item Development and validation of a robust, wind-agnostic path following controller for fixed-wing \acp{UAV}
    \item Experimental validation of the proposed architecture in simulation and a real-world flight test
\end{itemize}

\section{RELATED WORK}
Understanding how some bird species extract energy from the atmosphere has long been of interest~\cite{Rayleigh}. 
Optimization approaches have been used to understand the dynamic soaring process by finding the optimal trajectory to exploit energy from wind shear~\cite{zhao2004optimal, flanzer2012robust}. \cite{bousquet_optimal_2017} showed that the optimal dynamic soaring consists of large shallow arcs, which are consistent with observations. \cite{sachs_minimum_2005} estimates the minimum wind shear strength required for an Albatross to use dynamic soaring. 


Recent research has demonstrated interest in improving \ac{UAV} flight endurance by acquiring energy using thermal soaring~\cite{depenbusch_autosoar_2018, notter2023deep} or solar power~\cite{AtlantikSolar}. In contrast, utilizing dynamic soaring for \acp{UAV} has been relatively under-explored. \cite{lawrance2011path} solves a planning problem to find an optimal path using discrete maneuver libraries, where the wind field is updated using Gaussian process regression. Reinforcement learning approaches have also been used to generate a policy for optimizing energy gain through dynamic soaring~\cite{montella_reinforcement_2014, chung2015learning}. While there have been earlier attempts of dynamic soaring~\cite{bird2014closing, bronz_flight_2021}, these works do not demonstrate dynamic soaring on a real platform in realistic scenarios.

One of the main challenges in enabling dynamic soaring on a real vehicle is estimating the highly uncertain wind field. Estimating a global, time-averaged wind field is challenging because it primarily depends on direct, in-situ measurements, with wind gusts and sensor noise compounding the problem. In \cite{suys_autonomous_2023}, orographic soaring in a wind tunnel is demonstrated. \cite{bronz_simultaneous_2023} evaluates onboard wind estimates by comparing a ground-based Doppler lidar against wind estimates acquired by a quadrotor and an onboard multi-hole probe-based wind estimate. 
\cite{mckenna_online_2023} considers the uncertainty of the wind estimates by incorporating the estimation uncertainty online for the trajectory optimization problem. \cite{lawrance2011path} use Gaussian processes to model the uncertainty of the wind field.

\revision[
In this work, we use a robust trajectory optimization approach to ensure that the optimized trajectory is robust to wind estimation errors. For tracking, we use \iac{INDI} similar to~\cite{Tal}, where we show that the system is approximately differentially flat.]{
In this work, we take a different path by employing a \textit{robust} strategy in both planning and control.
A robust trajectory optimizer and \iac{INDI} tracking controller similar to~\cite{Tal} to ensure that the planned trajectories remain dynamically feasible, provide energy gain and can be tracked accurately in spite of significant, unknown perturbations in the wind field.
}

\section{PROBLEM FORMULATION}

\begin{figure}
    \centering
    \includegraphics[width=0.8\linewidth]{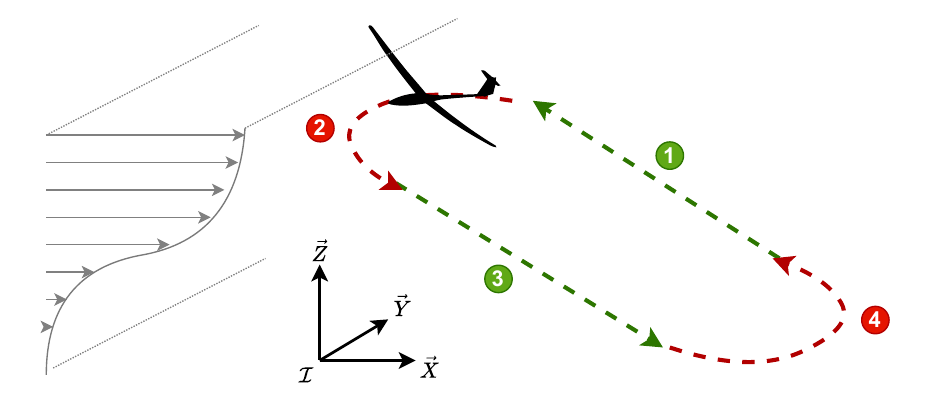}
    \caption{A simplified dynamic soaring cycle is described by four phases: upward wind climb (1) and downwind descent (3) are energy gain phases, while high altitude turn (2) and low altitude turn (4) describe energy loss phases.}
    \vspace{-6pt}
    \label{fig:ds_explanation}
\end{figure}
Dynamic soaring is a technique to extract energy from wind shear, which are present in boundary layers close to the ground, or in leeward regions of geographic features blocking the wind.
There are multiple types of DS cycles, e.g. wave-like or circular (Fig.~\ref{fig:ds_explanation}). The main problem is to find a DS cycle that can reliably extract energy from the wind shear.

\subsection{Wind Field Model}

We consider a horizontally-invariant, stationary wind field 
\begin{equation}
    \mathbf{w} = \sigma(z) \mathbf{w}_0 = \frac{1}{1 + e^{(-s (z - h))}}  \mathbf{w}_0
    \label{eq:wind_field}
\end{equation}
where $\mathbf{w}_0 = [w_{0,x},w_{0,y},0]^T$, $z$ is the altitude, and $h$ and $s$ are parameters determining the vertical location and thickness of the shear layer. 
The sigmoid distribution $\sigma(z)$ was adopted in previous work~\cite{bousquet_optimal_2017} to model a general shear layer.

\subsection{Aircraft Model}
\begin{figure}[t]
    \centering
    \includegraphics[width=0.3\textwidth]{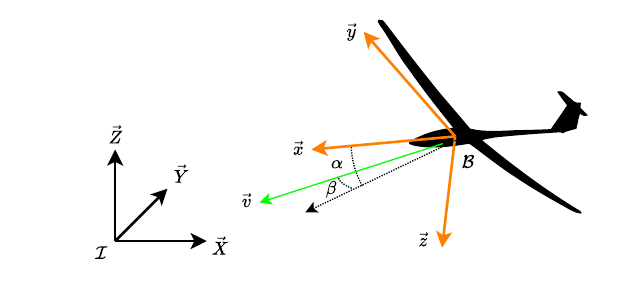}
    \caption{Frame conventions, $\mathcal{I}$ denotes the inertial \ac{ENU} frame, $\mathcal{B}$ denotes the body-fixed \ac{FRD} frame, $\vect{v}$ is the inertial velocity.}
    \label{fig:conventions}
\end{figure}

The fixed-wing aircraft is assumed to be a single rigid body with constant mass and inertia. All computations are performed in an earth-fixed \ac{ENU} reference frame  (Fig.~\ref{fig:conventions}).
For the modeling of the aircraft dynamics, a simple lumped sum model is employed for the aerodynamics in combination with the Newton-Euler equations for a rigid body \revision[]{with body-fixed frame $\mathcal{B}$}. The aerodynamic force $\vect{f}$ acting on the aircraft can be modeled as~\refequ{forces}, where $\rho$ is the air density, ${\vect{v}_{\text{air}}}$ is the air-relative velocity, $A_{\text{wing}}$ is the planform wing area,
$C_L, C_D$ are normalized lift and drag coefficients, and $\alpha$ is the air-relative \ac{AoA} of the wing~\cite{FlightMechanics}.
\begin{equation}\label{forces}
     \prescript{}{\mathcal{B}}{\vect{f}} = - \frac{1}{2} \rho ||{\vect{v}_{\text{air}}}||^2 A_{\text{wing}} \mathcal{R}_{\mathcal{B}\mathcal{A}}
     \begin{bsmallmatrix}
           C_{D} (\alpha) \\
           0 \\
           C_{L} (\alpha)
         \end{bsmallmatrix}
         + \prescript{}{\mathcal{B}}{\vect{f}_\text{ext}}
\end{equation}
$\mathcal{R}_{\mathcal{B}\mathcal{A}}$ is the rotation matrix to rotate the frame $\mathcal{B}$ around the body $y$-axis about the \ac{AoA} $\alpha$. All unmodelled forces are summarized as $\prescript{}{\mathcal{B}}{\vect{f}_\text{ext}}$.
Note that the lateral forces are ignored due to the zero side slip assumption. A linear relation between lift force, and a quadratic relation between drag force and \ac{AoA} is assumed as in \refequ{lift},~\refequ{drag}.
\begin{align}
    C_L (\alpha) &= C_{L0} + C_{L1}\alpha \,\label{lift}\\
    C_D (\alpha) &= C_{D0} + C_{D1}\alpha + C_{D2}\alpha^2\text{.}\label{drag}
\end{align}
Similar to the aerodynamic model employed in \cite{Tal}, the aerodynamic moment contribution due to control surface deflections is modeled as in~\refequ{moments},
\begin{equation}\label{moments}
     \prescript{}{\mathcal{B}}{\vect{m}} = \frac{1}{2} \rho ||{\vect{v}_{\text{air}}}||^2 
     \begin{bsmallmatrix}
           k_{\delta_{\text{ail}}} {\delta_{\text{ail}}}  \\
           k_{\delta_{\text{ele}}} {\delta_{\text{ele}}} \\
           k_{\delta_{\text{rud}}} {\delta_{\text{rud}}}
     \end{bsmallmatrix} + \prescript{}{\mathcal{B}}{\vect{m}_\text{ext}}
\end{equation}
where $k_{\delta_{\text{ail}}}$, $k_{\delta_{\text{ele}}}$, $k_{\delta_{\text{rud}}}$ are lumped parameters corresponding to aileron, elevator and rudder effectiveness and $\delta_{\text{ail}}$, $\delta_{\text{ele}}$, $\delta_{\text{rud}}$ are control surface deflections of aileron, elevator and rudder, respectively. $\prescript{}{\mathcal{B}}{\vect{m}_\text{ext}}$ summarizes the contribution of all unmodelled moments acting on the wing, fuselage and control surfaces. 
%

\begin{figure*}[t]
    \centering
    \includegraphics[width=\textwidth]{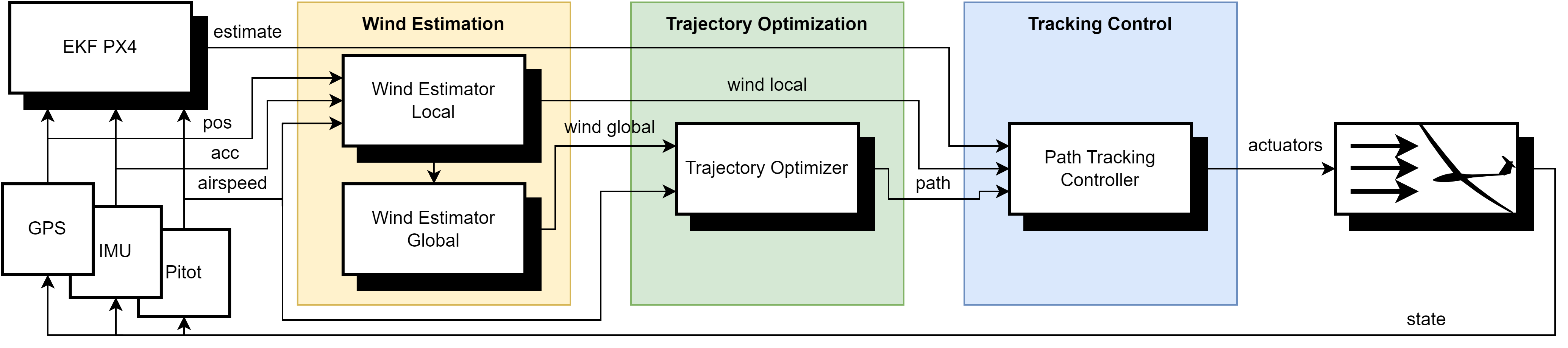}
    \caption{Proposed guidance and control architecture for dynamic soaring. The approach uses three separate modules for wind field estimation, trajectory optimization and path following control. Robustness to wind is achieved through two mechanisms. First, robust trajectory optimization ensures energy margins under wind variations. Second, path tracking control to minimize tracking error using only a local wind estimate and incremental feedback.}
    \label{fig:architecture}
\end{figure*}

\section{DYNAMIC SOARING FRAMEWORK}

The \revision[overview of the dynamic soaring framework is shown in \reffig
{fig:architecture}. It]{proposed dynamic soaring framework} consists of individual modules for the wind estimation,
trajectory optimization and tracking control\revision[]{~(\reffig
{fig:architecture})}. 
The modules are structured such that uncertain wind fields are either explicitly accounted for \textit{a priori} (robust path optimizer) or explicitly estimated (local wind estimator) and attenuated (path following controller).

\subsection{Wind Field Estimation}

\begin{figure}[t]
    \centering
    \includegraphics[width=\linewidth]{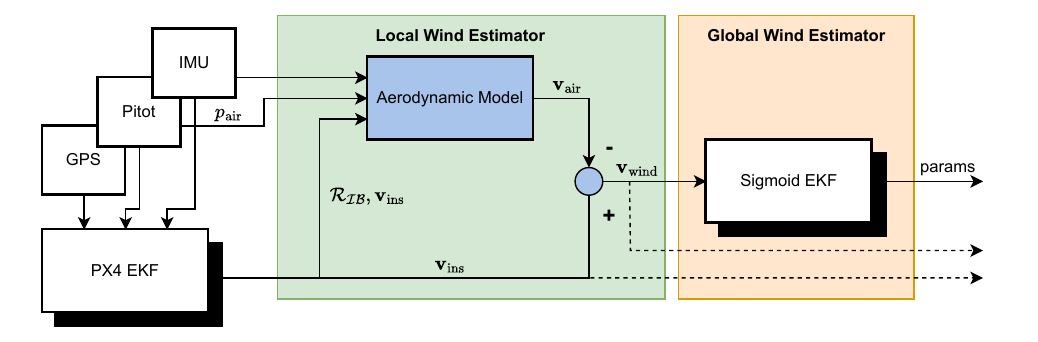}
    \caption{Block diagram of the wind estimation. Measurements from the pitot, IMU and GPS are fused to estimate the parameters of a sigmoidal wind field function.}
    \label{fig:wind_architecture}
\end{figure}

The wind estimation module consists of a local wind estimator and a global wind estimator~(\reffig{fig:wind_architecture}). The local wind estimator estimates the three-dimensional wind vector at the current position, and the global wind estimator fuses these local wind measurements to estimate the spatial wind field. This allows the local, instantaneous wind estimate to be used for control, while global wind estimates to be used for trajectory optimization. This effectively decouples tracking performance and gust rejection from a globally consistent wind estimate.

\subsubsection{Local Wind Estimator} The local wind estimator combines sensor measurements with a first-principles aerodynamic model and a wind triangle to produce pseudo measurements of the local, instantaneous wind vector $\vect{w}$. The air-relative velocity in the body frame $\prescript{}{\mathcal{B}}{\vect{v}}_\text{air}=[u,v,w]^T$ is approximated from direct airspeed measurements ($z_\text{TAS}$) and a polynomial fit of the inverted aerodynamic model.
\begin{equation}
    \begin{split}
        u &= z_\text{TAS}\\
        v &= \frac{-\prescript{}{\mathcal{B}}{\vect{f}}_y}{0.5 \cdot \rho \cdot (z_\text{CAS})^2 \cdot A_\text{wing}} \cdot \frac{u}{C_{B1}}\\
        w &= \left(\frac{-\prescript{}{\mathcal{B}}{\vect{f}}_z}{0.5 \cdot \rho \cdot (z_\text{CAS})^2 \cdot A_\text{wing}} - C_{A0}\right) \cdot \frac{u}{C_{A1}},
    \end{split}
    \label{wind estimation}
\end{equation}
with $C_{A0}$, $C_{A1}$, $C_{B1}$ being aerodynamic coefficients \revision[]{ , where the coefficients were determined from a nonlinear least-squares fit from flight data and are shown in Table \ref{tab:parameters}}. The aerodynamic forces $_\mathcal{B}\vect{f}_{y,z} = \vect{a}_{IMU,y,z}/m$ are derived from accelerometer readings. Further, $z_\text{TAS}$ is the true airspeed and $z_\text{CAS}$ is the calibrated airspeed.
Finally, the wind velocity vector $\vect{w}$ can be computed by applying a wind triangle computation with the estimated inertial velocity $\vect{v}_\text{ins}$:
\begin{equation}
    \vect{w} = \vect{v}_\text{ins} - \vect{v}_\text{air} \text{.}
\end{equation}

\subsubsection{Global Wind Estimator} The global wind estimator fuses local wind estimates to estimate the spatial distribution of the wind field. 
As the sigmoid function \eqref{eq:wind_field} is nonlinear in the parameters $\alpha$ and $h$, an \ac{EKF} makes use of local linearization of the observation dynamics. Therefore, the convergence guarantee of the linear Kalman Filter does not apply to this estimation problem\revision[~\cite{MPC}]{}. Also, the wind field is only partially observable (e.g. an agent taking local wind field measurements must cover some subspace of the 3D space to estimate the wind field). 

For robustness of the proposed estimation algorithm, the parametrization of the wind field is thus augmented by two additional parameters $b_x$ and $b_y$, representing the static wind components (bias terms):
\begin{equation}\label{wind field 2}
    w_x = w_{0,x} \cdot \sigma(z) + b_x\,\text{,}\quad w_y = w_{0,y} \cdot \sigma(z) + b_y \,\text{,}\quad  w_z = 0.
\end{equation}
The resulting state of the estimator is thus given by $\mathbf{x} = [w_{0,x},w_{0,y},b_x,b_y,s,h]^T$, and a trivial (\revision[]{identity plus }random walk) process model is assumed for the wind field parameters due to the stationarity assumption. \revision[]{We use identity matrices multiplied with $1e^{-6}$ and $4$ for the process and measurement covariances respectively and run the \ac{EKF} at \SI{10}{\hertz}.} The estimate of the global wind field $w_x,w_y,s,h$ is then used for trajectory optimization explained in the following subsection.

\subsection{Trajectory Optimization}%
Robust dynamic soaring trajectories for a given wind field and initial velocity of the vehicle are computed with trajectory optimization. 
To ensure sustained DS in a real world scenario, it is crucial to consider the uncertainty of the wind field. We therefore encode robustness point-wise with respect to wind field parameter variations in our trajectory optimization. 

We resort to a path-following approach, where the output of the trajectory optimizer is interpreted as a path. This spatial path\revision[, through different time parameterizations and states other than position, is mapped to a set of feasible trajectories.]{ is mapped to a set of feasible trajectories, through different time parameterizations and states other than position.} \revision[This allows to handle the infeasibility to track a trajectory due to the lack of thrust input.]{This allows to handle the infeasibility of trajectory tracking resulting from the lack of a thrust input.} In the following formulation, the set of trajectories $\mathcal{X}$ corresponds to a set of different wind fields $\mathcal{W}$. Other \revision[sources of uncertainty]{uncertainties}, such as \revision[w.r.t. ]{}aerodynamic coefficients, can \revision[easily ]{}be accounted for in a similar way.

To setup the trajectory optimization, we employ the kinematic system model from \cite{deittert2009engineless} with the state and input vector\revision[defined as]{}
\begin{equation}\label{kinematic states}
    \mathbf{x} = [x,y,z,V,\gamma,\psi]^T, \quad
    \mathbf{u} = [C_L,\mu]^T \text{,}
\end{equation}
where $x$, $y$, $z$ denotes the position of the aircraft in the inertial frame $\mathcal{I}$, $V$ is the airspeed, $\gamma$ is the glide path angle in the earth frame and $\psi$ is the aircraft heading. The control inputs to the model are given by the lift coefficient $C_L$ (controlled via angle of attack) and the bank angle $\mu$.

We approximate the set of all possible wind fields by sampling a finite number of discrete fields $\mathcal{W} = \{\vect{w}^0, \dots , \vect{w}^j, \dots , \vect{w}^M \}$ and their corresponding trajectories $\mathcal{X} = \{\{\mathbf{x}_i^0\}_{i=0}^{N}, \dots , \{\mathbf{x}_i^j\}_{i=0}^{N}, \dots , \{\mathbf{x}_i^M\}_{i=0}^{N} \}$. Since all elements of $\mathcal{X}$ are required to have the same path and the airspeed of the gliding \ac{UAV} is not controllable independently of the glide path, the trajectories in $\mathcal{X}$ require different time parameterizations and control inputs. We therefore introduce a set of time intervals $\mathcal{T} = \{\{\mathbf{t}_i^0\}_{i=0}^{N-1}, \dots , \{\mathbf{t}_i^j\}_{i=0}^{N-1}, \dots , \{\mathbf{t}_i^M\}_{i=0}^{N-1} \}$ and a set of control inputs $\mathcal{U} = \{\{\mathbf{u}_i^0\}_{i=0}^{N-1}, \dots , \{\mathbf{u}_i^j\}_{i=0}^{N-1}, \dots , \{\mathbf{u}_i^M\}_{i=0}^{N-1} \}$, where the subscript $i$ corresponds to the $i^{\text{th}}$ time step and the superscript $j$ corresponds to the $j^{\text{th}}$ element in $\mathcal{W}$. The resulting \ac{OCP} to compute point-wise robust paths is now given by \eqref{robust OCP_implementation_split}.
\ifthenelse{\boolean{show_revisions}}
{
    {\color{red}
    \begin{align}\label{robust OCP_OLD} 
        J^* (\mathbf{x},&\mathbf{u},\mathbf{t})  =  \max_{\mathbf{x}_{0},\{\mathbf{u}_i,\mathbf{t}_i\}_{i=0}^{N-1}} &  &  \min_{\{j\}_{j=0}^{M}} V_N^j\\
            \textrm{s.t.} \quad & \mathbf{x}_{i+1}^j = f_j(\mathbf{x}_i^j,\mathbf{u}_i^j,\mathbf{t}_i^j) & \forall j &\in  \{0,M\}, \forall i \in  \{0,N-1\}\nonumber\\
             & \mathbf{x}_i^j \in \mathcal{C_X} &  \forall j &\in  \{0,M\}, \forall i \in  \{0,N\}\nonumber\\
             & \mathbf{u}_i^j \in \mathcal{C_U} & \forall j &\in  \{0,M\}, \forall i \in  \{0,N-1\}\nonumber\\
             & \mathbf{x}_{0}^j \in \mathcal{C_X}_0 & \forall j &\in  \{0,M\}\nonumber\\
             & \mathbf{x}_{N}^j \in \mathcal{C_X}_N & \forall j &\in  \{0,M\}\nonumber\\
             & S \mathbf{x}_{i}^0 = S \mathbf{x}_{i}^j  & \forall j &\in  \{1,M\}, \forall i \in  \{0,N\}\nonumber
    \end{align}
    }
    
{
\color{blue}
\begin{subequations}\label{robust OCP_implementation_split}
\begin{align}
    J^* (\mathbf{x},&\mathbf{u},\mathbf{t}) = 
    \max_{
        \substack{
            \mathbf{x}_{0}, \Delta t, \{\mathbf{p}_i\}_{i=0}^{N}, \\
            \{\{\mathbf{u}_i^j,\mathbf{t}_i^j\}_{i=0}^{N-1}\}_{j=0}^{M}
        }
    } V_\text{min} \label{eq:objective}\\
    \textrm{s.t.} \quad 
    & V_\text{min} \leq \{V_N^j\}_{j=0}^{M}, \quad \forall j \in \{0,M\} \label{eq:vel}\\
    & \mathbf{x}_{i+1}^j = f_j(\mathbf{x}_i^j,\mathbf{u}_i^j,\mathbf{t}_i^j)\\ \nonumber
    & \hspace{6em} \forall j \in \{0,M\}, i \in \{0,N-1\} \label{eq:dyn}\\
    & S \mathbf{x}_{i}^j = \mathbf{p}_{i}, \quad \forall j \in \{0,M\}, i \in \{0,N\} \label{eq:coupling}\\
    & t_i^0 = \Delta t, \quad \forall i \in \{0,N\} \label{eq:time}\\
    & \mathbf{x}_i^j \in \mathcal{C_X}, \quad \forall j \in \{0,M\}, i \in \{0,N\} \label{eq:state_constr}\\
    & \mathbf{u}_i^j \in \mathcal{C_U}, \quad \forall j \in \{0,M\}, i \in \{0,N-1\} \label{eq:input_constr}\\
    & \mathbf{x}_{0}^j \in \mathcal{C_X}_0, \quad \mathbf{x}_{N}^j \in \mathcal{C_X}_N, \quad \forall j \in \{0,M\} \label{eq:init_constr}
\end{align}
\end{subequations}
}
}
{
\begin{subequations}\label{robust OCP_implementation_split}
\begin{align}
    J^* (\mathbf{x},&\mathbf{u},\mathbf{t}) = 
    \max_{
        \substack{
            \mathbf{x}_{0}, \Delta t, \{\mathbf{p}_i\}_{i=0}^{N}, \\
            \{\{\mathbf{u}_i^j,\mathbf{t}_i^j\}_{i=0}^{N-1}\}_{j=0}^{M}
        }
    }  V_\text{min} \label{eq:objective}\\
    \textrm{s.t.} \quad 
    & V_\text{min} \leq \{V_N^j\}_{j=0}^{M}, \quad \forall j \in \{0,M\} \label{eq:vel}\\
    & \mathbf{x}_{i+1}^j = f_j(\mathbf{x}_i^j,\mathbf{u}_i^j,\mathbf{t}_i^j) \label{eq:dyn} \\ 
    & \hspace{6em} \forall j \in \{0,M\}, i \in \{0,N-1\} \nonumber \\ 
    & S \mathbf{x}_{i}^j = \mathbf{p}_{i}, \quad \forall j \in \{0,M\}, i \in \{0,N\} \label{eq:coup}\\
    & t_i^0 = \Delta t, \quad \forall i \in \{0,N\} \label{eq:time}\\
    & \mathbf{x}_i^j \in \mathcal{C_X}, \quad \forall j \in \{0,M\}, i \in \{0,N\} \label{eq:state_constr}\\
    & \mathbf{u}_i^j \in \mathcal{C_U}, \quad \forall j \in \{0,M\}, i \in \{0,N-1\} \label{eq:input_constr}\\
    & \mathbf{x}_{0}^j \in \mathcal{C_X}_0, \quad \mathbf{x}_{N}^j \in \mathcal{C_X}_N, \quad \forall j \in \{0,M\} \label{eq:init_constr}
\end{align}
\end{subequations}
}
\revision[Note that $f_j(\cdot)$ denotes the system dynamics in the $j^{\text{th}}$ wind field. The common path constraint is enforced by the last three constraints. Note that we choose to maximize the worst-case terminal velocity $\forall \vect{w} \in \mathcal{W}$.]{
The above problem effectively maximizes the worst-case terminal velocity for all wind fields $\forall \vect{w} \in \mathcal{W}$ \eqref{eq:vel}, while the path constraint \eqref{eq:coup} with the selection matrix $S$ extracting position to ensure that all trajectories share a common path.
In \eqref{eq:dyn}, $f_j(\cdot)$ denotes the system dynamics in the $j^{\text{th}}$ wind field. The various constraint sets are specified in Table \ref{tab:constraints}.

\begin{figure}[t]
    \centering
    \includegraphics[width=0.5\textwidth]{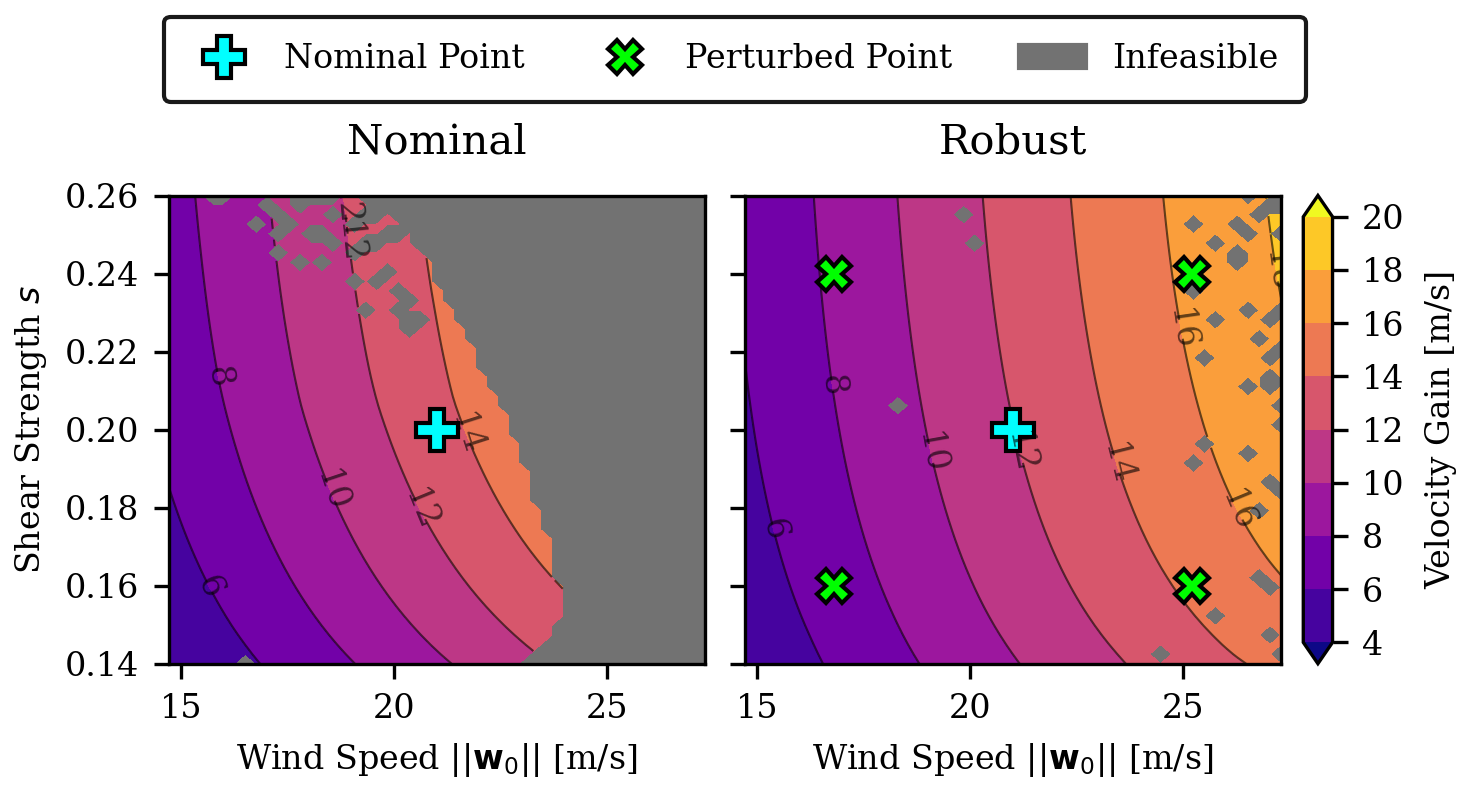}
    \caption{\revision[Robust energy gain paths, colored by the initial velocity at the lowest point of the path.]{Combined feasibility and sensitivity analysis of the robust paths. The robust paths considering 4 perturbed wind fields remain dynamically feasible for significantly larger perturbations in the encountered wind field.}}
    \label{fig:robust_trajectories}
\end{figure}
}
\ifthenelse{\boolean{show_revisions}}{
\begin{table}[]
\color{blue}
    \centering
    \caption{Constraint sets used in the optimization}
    \label{tab:constraints}
    \footnotesize 
    \renewcommand{\arraystretch}{0.9} 
    \begin{tabular}{l l l}
    \toprule
    \textbf{Set} & \textbf{Quantity} & \textbf{Definition / Bound} \\
    \midrule
    
    \multirow{3}{*}{$\mathcal{C_U}$} 
      & Lift $C_L$ & $[0, 1.0]$ \\
      & Bank $\mu$ & $[-\pi, 0]$ \\
      & Roll Rate & $|\Delta \mu| \leq 2.0 \cdot \Delta t$ \\
    \midrule
    
    \multirow{4}{*}{$\mathcal{C_X}$} 
      & Speed $v$ & $[10, 100]$ \\
      & Altitude $z$ & $[-50, 50], \ z_k \geq z_0$ \\
      & Heading $\psi$ & $\psi_{k+1} \leq \psi_k$ \\
      & Time step $\Delta t$ & $[0.05, 1.0]$ \\
    \midrule
    
    \multirow{2}{*}{$\mathcal{C_X}_0$} 
      & Pos. $\mathbf{p}_0$ & $[0, 0, z_0]^T, \ z_0 \in [-50, 0]$ \\
      & Angles $\gamma_0, \psi_0$ & $0$ \\
    \midrule
    
    \multirow{2}{*}{$\mathcal{C_X}_N$} 
      & State & $\mathbf{p}_N = \mathbf{p}_0, \ \gamma_N = \gamma_0$ \\
      & Heading $\psi_N$ & $\psi_0 - 2\pi$ \\
    
    \bottomrule
    \end{tabular}
\end{table}
}
{
\begin{table}[b]
    \centering
    \caption{Constraint sets used in the optimization}
    \label{tab:constraints}
    \footnotesize 
    \renewcommand{\arraystretch}{0.9} 
    \begin{tabular}{l l l}
    \toprule
    \textbf{Set} & \textbf{Quantity} & \textbf{Definition / Bound} \\
    \midrule
    
    \multirow{3}{*}{$\mathcal{C_U}$} 
      & Lift $C_L$ & $[0, 1.0]$ \\
      & Bank $\mu$ & $[-\pi, 0]$ \\
      & Roll Rate & $|\Delta \mu| \leq 2.0 \cdot \Delta t$ \\
    \midrule
    
    \multirow{4}{*}{$\mathcal{C_X}$} 
      & Speed $v$ & $[10, 100]$ \\
      & Altitude $z$ & $[-50, 50], \ z_k \geq z_0$ \\
      & Heading $\psi$ & $\psi_{k+1} \leq \psi_k$ \\
      & Time step $\Delta t$ & $[0.05, 1.0]$ \\
    \midrule
    
    \multirow{2}{*}{$\mathcal{C_X}_0$} 
      & Pos. $\mathbf{p}_0$ & $[0, 0, z_0]^T, \ z_0 \in [-50, 0]$ \\
      & Angles $\gamma_0, \psi_0$ & $0$ \\
    \midrule
    
    \multirow{2}{*}{$\mathcal{C_X}_N$} 
      & State & $\mathbf{p}_N = \mathbf{p}_0, \ \gamma_N = \gamma_0$ \\
      & Heading $\psi_N$ & $\psi_0 - 2\pi$ \\
    
    \bottomrule
    \end{tabular}
\end{table}
}
\revision[In
practice, an equivalent but better-structured OCP is used to
pre-compute robust paths]{
We use the \ac{OCP} \eqref{robust OCP_implementation_split} to pre-compute robust paths} for a (uniform, grid) sampling of \revision[]{nominal} wind fields \revision[]{with 4 perturbed wind fields each ($M=5$). The increased robustness under uncertainty is shown in Fig. \ref{fig:robust_trajectories}, where some efficiency is traded for robust feasibility of a \textit{fixed path} under constraints (Table \ref{tab:constraints}) and uncertainty in the wind field. Note that this differs from the feasibility of the constrained, nominal problem under a known wind field.} \revision[and for initial velocities offline.]{We use the IPOPT solver with implicit midpoint integration and multiple shooting in our implementation to compute the robust paths for varying initial velocities in an offline step.} Only the common reference path of the optimal solution of \eqref{robust OCP_implementation_split} is stored and then passed to the tracking controller. For differentiability of the path kinematics, we use a B-spline representation of the path with basis functions $\Phi = \{\phi_0(\tau), \phi_1(\tau), \hdots, \phi_{15}(\tau)\}$, where $\tau \in [0,1]$ is the normalized time variable during the dynamic soaring cycle. The \revision[basis]{basis\footnote{The chosen basis is $C^\infty$ continuous $\forall t \in (0,1)$, but only $C^0$ continuous at $t=0 \rightarrow t=1$. A Bernstein basis is an alternative choice here.}} is given by 
\begin{equation}
    \phi_i(\tau) = \begin{cases}
    1 &  i=0\\
    \sin(\tau\pi)\cdot \exp(-30((\tau-\frac{i}{16})^2)) & \forall i \in \{1,15\}\\
    \end{cases}\nonumber
\end{equation}
and the linear coefficients are determined by linear regression on a set of control points on the path. Therefore, each dynamic soaring path is represented by a set of 16 coefficients per dimension, 48 coefficients in total.

\subsection{Path Following Control}
\revision[
Once a path has been generated, the vehicle tracks it using a dynamic-inversion based controller.
Dynamic inversion requires analytic invertibility and differential flatness of the model. However, for a fixed-wing gliding aircraft, tracking of any arbitrary trajectory in state space cannot be achieved due to a lack of degrees of freedom to generate thrust and direct yaw control. 
Therefore, we employ a projection of the flat outputs onto a feasible subspace, which can be tracked by the aircraft. This projection allows us to make use of differential flatness based control methodologies for \emph{path following} without encountering infeasible configurations.]{
Once a path is generated, the vehicle tracks it using a dynamic-inversion based controller. However, for a gliding aircraft, full differential flatness is unattainable due to the lack of thrust and direct yaw authority. We therefore project the flat outputs onto a feasible subspace, allowing a \textit{path following} approach without encountering infeasible configurations.
}

\begin{figure}[t]
    \centering
    \includegraphics[width=\linewidth]{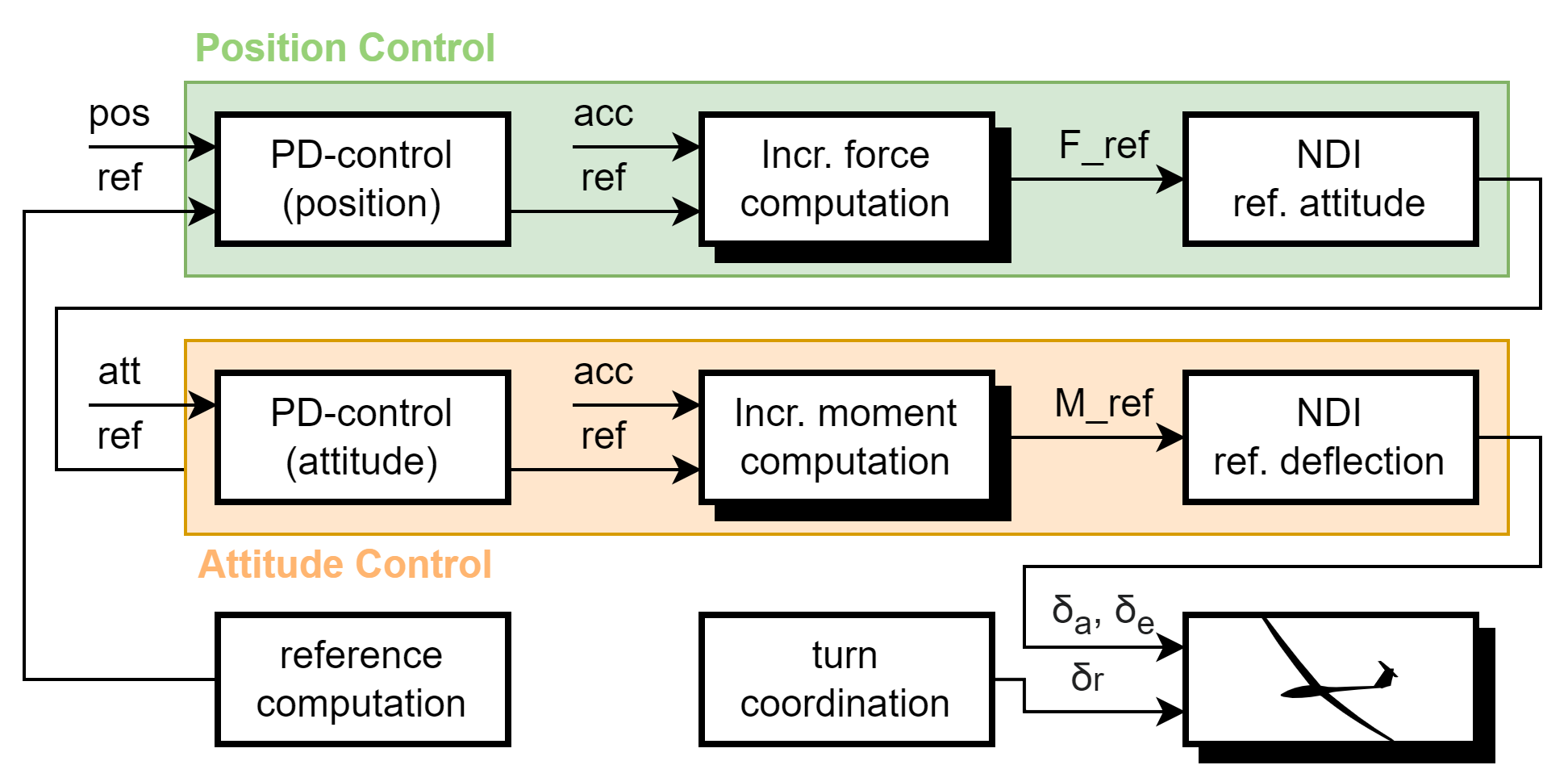}
    \caption{Proposed \ac{INDI} control architecture for path following control.}
    \label{fig:control_architecture}
\end{figure}

\subsubsection{Position Dynamics}
We construct projected flatness relations for tracking a position reference $\xi = \vect{r}_\text{ref}(t)$.
The required force ${\vect{f}}_\text{ref}$ to fulfill $\ddot{\vect{r}} = \ddot{\vect{r}}_\text{ref}$ is\revision[, in general, obtained by simply plugging  $\ddot{\vect{r}}_\text{ref}$ into the]{ obtained from the} Newton-Euler equation
\begin{equation}\label{force ref}
    \prescript{}{\mathcal{B}}{\vect{f}}_\text{ref} = m \cdot \mathcal{R}_{\mathcal{BI}}(
    \prescript{}{{\mathcal{I}}}{\ddot {\vect{r}}_\text{ref}}-  \prescript{}{\mathcal{I}}{\vect{g}}) \text{.}
\end{equation}
\revision[
For dynamic inversion, we must equate \eqref{forces} with \eqref{force ref}, which, however, is infeasible due to the coupling between lift and drag. Instead, we only consider the perpendicular component to the path, which is necessary to track lateral kinematics. We decompose the required net force into its lift and drag components, i.e.]{
Direct inversion via \eqref{forces} is infeasible due to lift-drag coupling. Instead, we project the requirement onto the path-perpendicular vector to satisfy lateral kinematics, decomposing the net force as
}
\begin{equation}\label{lift_projected}
    \prescript{}{\mathcal{B}}{\vect{f}}_\text{ref}^\text{lift} = \prescript{}{\mathcal{B}}{\vect{f}}_\text{ref} - \prescript{}{\mathcal{B}}{\vect{f}}_\text{ref}^\text{drag}
    \text{,} \quad 
    \prescript{}{\mathcal{B}}{\vect{f}}_\text{ref}^\text{drag} = 
    \frac{\prescript{}{\mathcal{B}}{\vect{f}}_\text{ref}\cdot{\prescript{}{\mathcal{B}}{\vect{v}}_{\text{air}}^\text{ref}}}{||{\prescript{}{\mathcal{B}}{\vect{v}}_{\text{air}}^\text{ref}}||^2} \cdot {\prescript{}{\mathcal{B}}{\vect{v}}_{\text{air}}^\text{ref}}
    \text{,}
\end{equation}
where $\vect{v}_\text{air}^\text{ref} = \dot{\vect{r}}_\text{ref} - \vect{w}$ for the reference velocity $\dot{\vect{r}}_\text{ref} $ and the current local wind vector $\vect{w}$.
The lift force $\prescript{}{\mathcal{B}}{\vect{f}}_\text{ref}^\text{lift}$ neglects any force components that are parallel to the reference trajectory in the air-relative frame.  
The lift force can now be used to perform the dynamic inversion, i.e. to find the corresponding attitude. The required drag force is neglected, constituting the mentioned projection.
By considering the force model \eqref{forces} with a zero side slip assumption, we can equate this with equation \eqref{lift_projected} and set $\mathbf{F}_\text{ext} = 0$ and $\prescript{}{\mathcal{B}}{\vect{f}}_\text{ref}^\text{lift} = \prescript{}{\mathcal{B}}{\vect{f}}$ and obtain
\begin{equation}\label{lift constraint}
    \prescript{}{\mathcal{I}}{\vect{f}}_\text{ref}^\text{lift}
    =  - \frac{1}{2} \rho ||{\vect{v}_{\text{air}}}||^2 A_{\text{wing}} \mathcal{R}_{\mathcal{I}\mathcal{B}} \mathcal{R}_{\mathcal{B}\mathcal{A}}
     \begin{bsmallmatrix}
       C_{D} (\alpha) \\
       0 \\
       C_{L} (\alpha)
     \end{bsmallmatrix}
    \text{.}
\end{equation}
Considering the lift in the air-relative frame, we construct the required attitude to achieve the reference lift by a two-stage rotation. First, we construct a zero-\ac{AoA} frame $\mathcal{B}_0$ with forward $\mathcal{B}_{0,x}$ aligned with the airspeed $\mathbf{v}_{air}$ to satisfy the zero side slip and $\mathcal{B}_{0,z}$ aligned with the negative lift force as 
\begin{equation}
    \mathcal{R}_{\mathcal{I}\mathcal{B}_0} =
    \left[
    \begin{array}{c|c|c}
       \frac{{\prescript{}{\mathcal{I}}{\vect{v}}_{\text{air}}^\text{ref}}}{||{\prescript{}{\mathcal{I}}{\vect{v}}_{\text{air}}^\text{ref}}||}
       &
       \frac{\prescript{}{\mathcal{I}}{\vect{f}}_\text{ref}^\text{lift} \times {\prescript{}{\mathcal{I}}{\vect{v}}_{\text{air}}^\text{ref}}}{||\prescript{}{\mathcal{I}}{\vect{f}}_\text{ref}^\text{lift}|| \cdot ||{\prescript{}{\mathcal{I}}{\vect{v}}_{\text{air}}^\text{ref}}||}
       &
       \frac{\prescript{}{\mathcal{I}}{\vect{f}}_\text{ref}^\text{lift}}{||\prescript{}{\mathcal{I}}{\vect{f}}_\text{ref}^\text{lift}||}
    \end{array}\right] \text{.}
\end{equation}
Then, we additionally rotate the frame $\mathcal{B}_0$ around the body $y$-axis (effectively setting angle of attack) to satisfy Eq.~\eqref{lift constraint}
\begin{equation}
    \mathcal{R}_{{\alpha}} = \begin{bsmallmatrix}
      \cos (\alpha) & 0 & -\sin (\alpha)\\
      0 & 1 & 0\\
      \sin (\alpha) & 0 & \cos (\alpha)
    \end{bsmallmatrix} \text{,}
\end{equation}
where we obtain $\alpha$ from \eqref{lift} and \eqref{lift constraint} as
\begin{equation}
    \alpha = \frac{1}{C_{L1}} \left (\frac{2\cdot||{\vect{f}}_\text{ref}^\text{lift}||}{\rho ||\vect{v}_\text{air}||^2 A_\text{wing}} - C_{L0} \right) \text{.}
\end{equation}

The attitude reference that satisfies lift and zero side slip constraints in \eqref{lift constraint} can thus be constructed as
\begin{equation}\label{position inversion}
    \mathcal{R}_{\mathcal{I}\mathcal{B}}^\text{ref} =  \mathcal{R}_{\mathcal{I}\mathcal{B}_0} \mathcal{R}_{{\alpha}} \text{.}
\end{equation}

\subsubsection{Attitude Dynamics}
In the second step, we construct flatness relations for the output $\xi = \mathcal{R}_{\mathcal{I}\mathcal{B}}$ of the angular subsystem. For this, consider an attitude reference given as 
\begin{equation}
    \xi = \mathcal{R}_{\mathcal{I}\mathcal{B}}(t)
\end{equation}
where the attitude trajectory also includes information about temporal derivatives. The required moment to fulfill the kinematic constraint $\dot{\boldsymbol{\omega}} = \dot{\boldsymbol{\omega}}_\text{ref}$ is obtained by plugging $\dot{\boldsymbol{\omega}}_\text{ref}$ into the Newton-Euler equations:
\begin{equation}
    \prescript{}{\mathcal{B}}{\vect{m}_\text{ref}}  =  \prescript{}{B}{\vect{I}} \prescript{}{\mathcal{B}}{\dot {\boldsymbol{\omega}}_\text{ref}}  + \prescript{}{\mathcal{B}}{\boldsymbol{\omega}} \times (\prescript{}{\mathcal{B}}{\vect{I}} \prescript{}{\mathcal{B}}{\boldsymbol{\omega}}) \text{.}
\end{equation}
By considering the moment model \eqref{moments}, we can now substitute $\prescript{}{\mathcal{B}}{\vect{m}_\text{ref}}$ and set $\mathbf{M}_\text{ext} = 0$ to obtain
\begin{equation}\label{flatness moments}
     \prescript{}{B}{\vect{I}} \prescript{}{\mathcal{B}}{\dot {\boldsymbol{\omega}}_\text{ref}}  + \prescript{}{\mathcal{B}}{\boldsymbol{\omega}} \times (\prescript{}{\mathcal{B}}{\vect{I}} \prescript{}{\mathcal{B}}{\boldsymbol{\omega}}) = \frac{1}{2} \rho ||{\vect{v}_{\text{air}}}|| 
     \begin{bsmallmatrix}
           k_{\delta_{\text{ail}}} {\delta_{\text{ail}}}  \\
           k_{\delta_{\text{ele}}} {\delta_{\text{ele}}} \\
           k_{\delta_{\text{rud}}} {\delta_{\text{rud}}}
     \end{bsmallmatrix} \text{.}
\end{equation}
While differential flatness should be achieved with the model above, the unmodelled restoring moment due to sideslip limits control action around the body $z$-axis to a minimum. This becomes clear as an increment in yaw angle is achieved by coordinating all actuator deflections $\delta_{\text{ail}}$, $\delta_{\text{ele}}$ and $\delta_{\text{rud}}$. By considering the proper flatness relation for the model above, an increment in body $z$-axis would be achieved by an increment in $\delta_{\text{rud}}$, which is not practically achievable. This is resolved by projecting the flatness relations onto the roll and pitch axes, and using the rudder only to counter the estimated side slip.

The local wind estimate and current reference path are used to compute actuator deflections by considering the flatness relations (Fig.~\ref{fig:architecture}). To allow compatibility of the path following approach with the flatness relations, the reference kinematic (acceleration command ${\ddot {\vect{r}}_\text{ref}}$) is constructed from the closest point on the reference path $\vect{r}_\text{ref}$ by local differentiation, rescaled to the current inertial speed.

To compensate for modeling errors and disturbances, the sensor-based control approach \ac{INDI} is chosen to control both the position and attitude of the aircraft. 
We choose to employ a cascaded control approach inspired by \cite{Tal} for position and attitude control of the gliding aircraft. The cascaded control architecture is shown in Fig.~\ref{fig:control_architecture}.

\subsubsection{Position Control}
In the position controller, the position, velocity, and acceleration reference and related errors are used to compute an attitude reference for the attitude controller. First, a linear PD control law is applied to the position and velocity errors to obtain an acceleration command for the actual system. Similar to \cite{Tal}, the control law can be written as
\begin{equation}\label{position PD control}
    \ddot{\vect{r}}_\text{c} = k_\text{P}^\text{pos}(\vect{r}_\text{ref}-\vect{r}) +
    k_\text{D}^\text{pos}(\dot{\vect{r}}_\text{ref}-\dot{\vect{r}}) +
    \ddot{\vect{r}}_\text{ref} \text{,}
\end{equation}
where ${\vect{r}}$ and $\dot{\vect{r}}$ are the current position and velocity, respectively. The references ${\vect{r}}_\text{ref}$, $\dot{\vect{r}}_\text{ref}$, $\ddot{\vect{r}}_\text{ref}$ are taken from the kinematics of the closest point on the reference path.
$k_\text{P}^\text{pos}$ and $k_\text{D}^\text{pos}$ are the proportional and derivative control gains, where $k_\text{D}^\text{pos}$ is selected such that the corresponding linear system is critically damped. 

Instead of directly allocating the \emph{net} control input (attitude reference) from $\ddot{\vect{r}}_\text{c}$ and the inversion of a likely erroneous model (incl. unknown disturbances), the employed incremental approach calculates force, and correspondingly attitude, \emph{increments} to compensate for the current acceleration tracking error. This effectively leverages acceleration measurements to observe, and cancel, unknown model components and disturbances and thereby leads to mentioned path following robustness. The force increment $\Delta \mathbf{F}$ is then added to the current force model to generate a force command $\mathbf{F}_\text{c}$
\begin{equation}
    \mathbf{F}_\text{c} = \underbrace{m(\ddot{\vect{r}}_\text{c}-\ddot{\vect{r}}^\text{lpf})}_{\Delta \mathbf{F}} + \mathbf{F}_\text{model}^\text{lpf} \text{,}
\end{equation}
where $m$ is the mass, $ \mathbf{F}_\text{model}$ is the current force model \eqref{forces} using the current state with $ \mathbf{F}_\text{ext}=0$ and the superscript `lpf' indicates the low-pass filtered signal. The low-pass filter is applied to attenuate measurement noise, introduced by the acceleration feedback $\ddot{\vect{r}}$, in the force command. This incremental structure allows the controller to add force increments until the system tracks the acceleration command given by the linear control structure in \eqref{position PD control}. In a third step, the force command is used to compute the attitude reference $\vect{q}_\text{c}$ using the flatness relation shown in \eqref{position inversion} and changing the representation to the corresponding unit quaternion representation.
\subsubsection{Attitude Control}
In the attitude controller, the reference values for the attitude $\vect{q}_\text{ref}$ and its temporal derivatives ${\boldsymbol{\omega}}_\text{ref}$ and $\dot{\boldsymbol{\omega}}_\text{ref}$ and the related errors are used to compute control surface deflections for the system. The structure of the attitude controller is identical to the structure shown in \cite{Tal}. However, the dynamic inversion is performed differently due to the zero sideslip assumption. In a first step, the error angle vector $\boldsymbol{\zeta}$ is obtained as 
\begin{equation}
    \boldsymbol{\zeta} = \frac{2 \arccos({q}_\text{e}^w)}{1-({q}_\text{e}^w)^2}
    \begin{bmatrix}
    {q}_\text{e}^x \quad {q}_\text{e}^y \quad {q}_\text{e}^z
    \end{bmatrix}^T
    \text{,} \quad \vect{q}_\text{e} = \vect{q}^{-1} \circ \vect{q}_\text{c}
\end{equation}
where 
$\vect{q}_\text{e} = \begin{bsmallmatrix}
    {q}_\text{e}^w \quad{q}_\text{e}^x \quad {q}_\text{e}^y \quad {q}_\text{e}^z
    \end{bsmallmatrix}^T$
is the attitude error quaternion and $\vect{q}$ is the current attitude quaternion. The feed-forward terms ${\boldsymbol{\omega}}_\text{ref}$ and $\dot{\boldsymbol{\omega}}_\text{ref}$ are derived from the on-path (closest point) target attitude required to follow the path with the current inertial speed and wind. The error angle vector $\vect{\zeta}$ is then used in a linear PD control law, to obtain an angular acceleration command for the system. The linear control law is written as
\begin{equation}\label{attitude PD control}
    \prescript{}{\mathcal{B}}{\dot{\boldsymbol{\omega}}}_\text{c} = \mathbf{K}_\text{P}^\text{att}(\boldsymbol{\zeta}) +
    \mathbf{K}_\text{D}^\text{att}(\prescript{}{\mathcal{B}}{\boldsymbol{\omega}}_\text{ref}-\prescript{}{\mathcal{B}}{\boldsymbol{\omega}}) +
    \prescript{}{\mathcal{B}}{\dot{\boldsymbol{\omega}}}_\text{ref} \text{,}
\end{equation}
where $\mathbf{K}_\text{P}^\text{att}$ and $\mathbf{K}_\text{P}^\text{att}$ are diagonal control gain matrices corresponding to the proportional and derivative terms, respectively. \revision[The diagonal entries of these matrices are selected according to the control authority available around the respective body axis. ]{}Analogously to the position controller,\revision[ in a second step,]{} the moment increment $\Delta \mathbf{M}$ is computed and added to the current modeled moment to create a moment command $\mathbf{M}_\text{c}$
\begin{equation}
    \mathbf{M}_\text{c} = \underbrace{\mathbf{I}(\dot{\boldsymbol{\omega}}_\text{c}-\dot{\boldsymbol{\omega}}^\text{lpf})}_{\Delta \mathbf{M}} + \mathbf{M}_\text{model}^\text{lpf} \text{,}
\end{equation}
where $\mathbf{M}_\text{model}$ is the current moment computed using the model in \refequ{moments}. Also, it is assumed that the angular term is slowly varying and \revision[is ]{}identical to its low-pass filtered counterpart \cite{Tal}. \revision[To compute $\mathbf{M}_\text{model}$, the current actuator deflection needs to be known. Howver, due to a lack of position feedback, we approximate the servo as first order system and estimate the deflection based on a corresponding low-pass filter applied to the actuator commands. ]{To obtain an estimate of $\mathbf{M}_\text{model}$, we approximate the servo as first order system and estimate the deflection based on a corresponding low-pass filter applied to the actuator commands. }Finally, the moment command is used to compute actuator deflection references for ailerons and elevator using the flatness relation \eqref{flatness moments}. 

Note that the rudder deflection is allocated independently to \revision[maintain]{enforce} zero side-slip. This is achieved through \revision[a ]{}\emph{turn coordination}\revision[. Based on the current speed and lateral acceleration readings from the IMU, it computes the required body-$z$ angular rate $r^*$ to maintain zero side-slip.]{, where the required body-$z$ angular rate $r^*$ to maintain zero side-slip is computed based on the current speed and lateral acceleration.}
The \revision[ideal]{desired} yaw rate $r^*$ is then passed \revision[through an airspeed scaling and fed ]{}to the rudder, together with a linear feedback term
\begin{equation}
    \delta_\text{rud} = k_\text{FF} \cdot r^* + k_\text{P} \cdot (r^* - r) ,
\end{equation}
where $k_\text{FF}$ and $k_\text{P}$ are the airspeed-scaled ($\propto z_{CAS}^{-2}$) feedforward and proportional control gains, respectively.
\begin{figure*}[t]
    \centering
    \includegraphics[width=\linewidth]{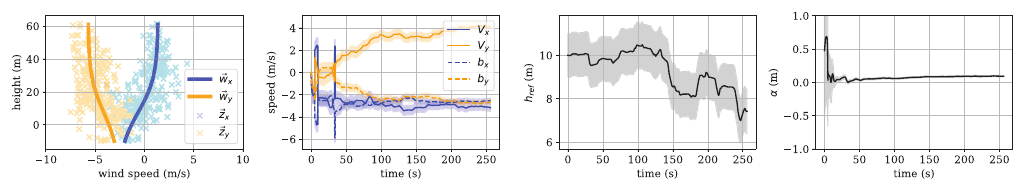}
    \begin{subfigure}[t]{0.24\linewidth}
    \vspace{-20pt}
    \caption{Estimated wind profile}
    \end{subfigure}
    \begin{subfigure}[t]{0.24\linewidth}
    \vspace{-20pt}
    \caption{Wind shear strength}
    \end{subfigure}
    \begin{subfigure}[t]{0.24\linewidth}
    \vspace{-20pt}
    \caption{Wind shear height}
    \end{subfigure}
    \begin{subfigure}[t]{0.24\linewidth}
    \vspace{-20pt}
    \caption{Sigmoidal steepness}
    \end{subfigure}
    \caption{Visualization of global wind estimator. a) Estimated wind shear profile. Convergence of wind shear parameters b) wind shear strength c) wind shear height d) sigmoidal parameter $\alpha$ shown as mean (solid line) and variance (colored).}
    \vspace{-10pt}
    \label{fig:shear_estimator_chasseral}
\end{figure*}
\section{EXPERIMENTAL RESULTS}
Validating \revision[a dynamic soaring framework]{the proposed approach} across a broad range of wind-conditions is \revision[practically challenging, due to limited flight testing time and often hard-to-predict wind conditions for planning effective field trips]{challenging, due to the stochastic nature of wind}. \revision[With this work]{Therefore}, we \revision[thus ]{}focus on evaluating \revision[the ]{}components of the framework individually, but stress important aspects for successful\revision[, integrated]{} real-world dynamic soaring. First, \revision[(robust) ]{}performance under a controllable range of wind-conditions and estimation errors is best assessed in simulation. Second, to ensure that the simulation results transfer to reality, we \revision[need to validate the wind-estimator and ]{}demonstrate that the system can i) closely follow paths as considered in the simulation and ii) that it exhibits a similar energy evolution\revision[, i.e.]{} one \revision[]{h}as modeled and optimized for. \revision[This step-by-step evaluation is presented in the following.]{}

\subsection{Simulation}
\begin{figure}
    \centering
    \includegraphics[width=0.95\linewidth]{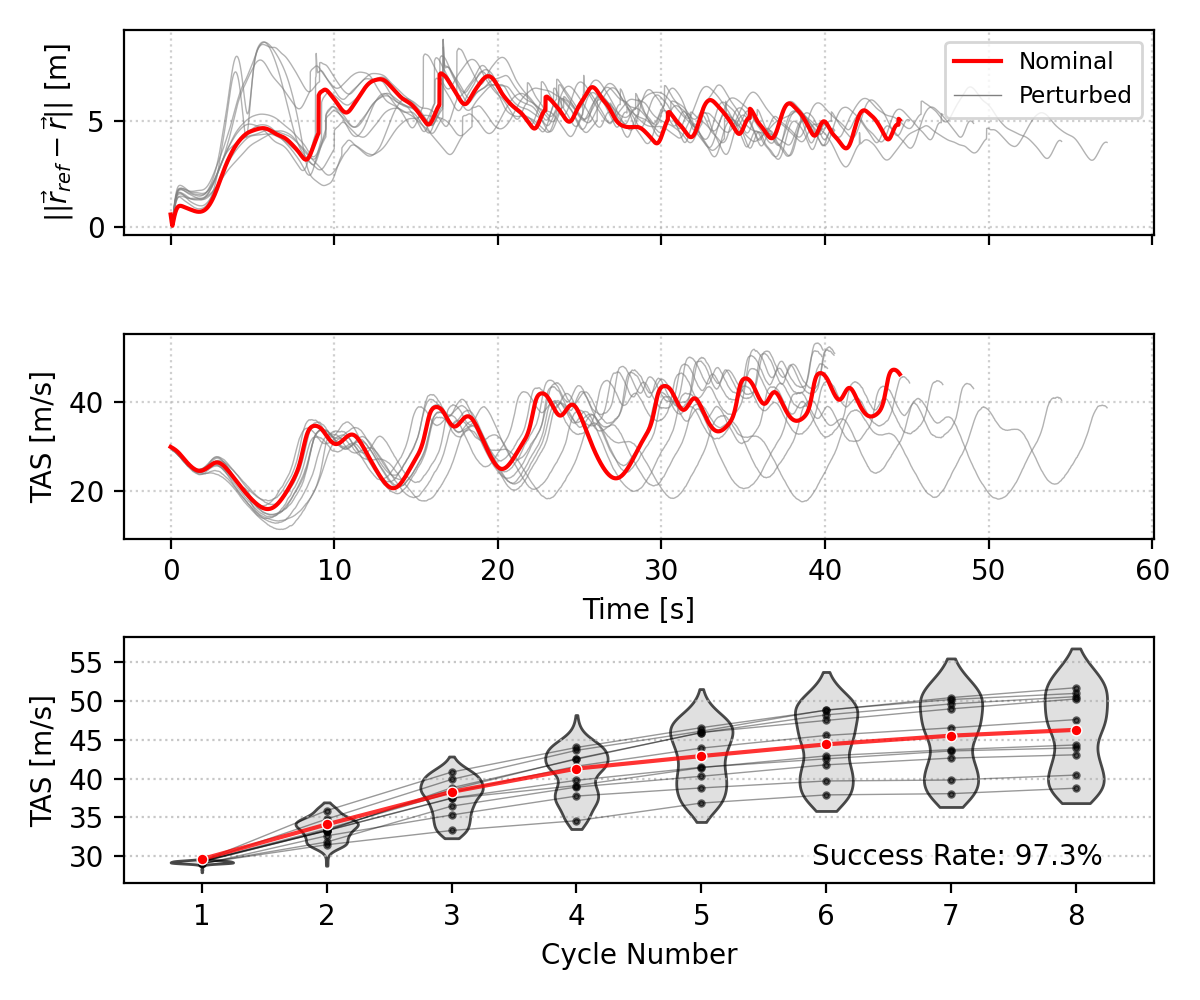}
    \vspace{-4pt}
    \caption{\revision[Monte-Carlo simulation of dynamic soaring with randomized wind fields. For an estimated, nominal wind field 
    (${w}_{max}=$~\SI{12}{\meter\per\second}, $s=$~\SI{0.4}{\meter}, $h_{ref}=$~\SI{0}{\meter}), the actual wind field parameters are perturbed by random additive Gaussian noise ($\sigma_{{w}_{max}}=$~\SI{1}{\meter\per\second}, $\sigma_{s}=$~\SI{0.05}{\meter}, $\sigma_{h_{ref}}=$~\SI{1}{\meter})
    and simulated for 20 such perturbed parameter sets. The discontinuous jumps in tracking error are due to reference switching at the end of a trajectory due to increased airspeed.]{Simulation of dynamic soaring in perturbed wind fields with overlayed turbulence from a Dryden model. The UAV follows robust soaring paths optimized for a nominal wind field with 
    ($||\mathbf{w}_0||=$~\SI{12}{\meter\per\second}, $s=$~\SI{0.3}{\meter}, $h_{ref}=$~\SI{0}{\meter}) and 4 perturbed wind fields with $25\%$ relative perturbations on $||\mathbf{w}_0||$ and $s$ each. In the simulation, the wind field parameters are perturbed by random additive, uniform noise with $25\%$ relative perturbation radius on $||\mathbf{w}_0||$ and $s$. The simulation is performed for $300$ runs (only $10$ shown for visual clarity).}}
    \label{fig:simulation}
\end{figure}

We simulate the closed-loop behavior of the trajectory optimizer and the path-tracking controller under wind estimation errors to evaluate the robustness of the pipeline. \revision[For this purpose, w]{W}e consider a nominal wind field ($\bar{w}_{max}=$~\SI{12}{\meter\per\second}, \revision[$\bar{s}=$~\SI{0.5}{\meter}]{$\bar{s}=$~\SI{0.3}{\meter}}, $\bar{h}_{ref}=$~\SI{0}{\meter}) which is used by the trajectory optimizer \revision[]{ alongside 4 perturbed wind fields ($M=5$) with a $25\%$ relative perturbation of the parameters $\bar{w}_{max}$ and $\bar{s}$}. 
The actual wind field parameters that are used in simulation are sampled from \revision[normal distributions (i.e. $w_{max} \sim \mathcal{N}\left(\bar{w}_{max}, \sigma_{w_{max}}^2\right)$) with $\sigma_{w_{max}}=$~\SI{1}{\meter\per\second}, $\sigma_{s}=$~\SI{0.05}{\meter}, $\sigma_{h_{ref}}=$~\SI{1}{\meter}. 
The closed loop system is simulated for \SI{60}{\second} per wind-field sample, i.e. the wind-field parameters are held constant during each rollout. Note that, while the reference path being tracked has been optimized based on the nominal wind-field, the tracking controller has access to noisy estimates of the actual, sampled wind field. This reflects the proposed use of a global (slower) and local (faster) wind-estimator. 
An example of the resulting responses can be seen in Fig.~\ref{fig:simulation}. The simulation results indicate that the proposed architecture can achieve robust dynamic soaring under unknown, static deviations in the wind field.]{a uniform distribution with a $25\%$ relative perturbation radius of $\bar{w}_{max}$ and $\bar{s}$. We use the Dryden model to add turbulence the mean wind field. Dryden model parameters $L$ and $\sigma$ were derived by fitting the model to the residual error signal from subtracting a Gaussian Process-modeled mean wind field from the flight data in Fig. \ref{fig:position_tracking_path1} (b). To further account for the worst-case turbulence, we increase the standard deviation to over-approximate 99.99\%, obtaining $L_u=\SI{36}{\meter}$, $L_v=\SI{20}{\meter}$, $L_w=\SI{10}{\meter}$, $\sigma_u=\SI{0.96}{\meter\per\second}$, $\sigma_v=\SI{0.84}{\meter\per\second}$, $\sigma_w=\SI{0.57}{\meter\per\second}$. We further add a time-varying, random estimation bias of \SI{1}{\meter} to the local wind estimate. During each rollout, the perturbed wind-field parameters are held constant and overlaid with the turbulence, while the global wind field for trajectory selection is kept at the nominal values. Trajectory switching due to changes in initial speed takes place at the end of a path segment. Note that, while the reference path being tracked has been optimized based on the nominal wind-field, the tracking controller has access to the actual, sampled wind field. This reflects the proposed use of a global and local wind-estimator. Robustness was evaluated over 8-cycle simulations, with success being completion without airspeed dropping below the \SI{10}{\meter\per\second}
  stall/tracking limit. Despite wind perturbations and modeling uncertainties, the architecture achieved a 97.3\% success rate and shows continuous increase of air speed in Fig. \ref{fig:simulation}.}
 
\subsection{Flight Experiments}
Hardware experiments were conducted with a \revision[powered (folding-prop)]{} model glider \revision[aircraft]{Multiplex EasyGlider} with a wingspan of \SI{1.8}{m} and a mass of \SI{1.35}{kg}. Wind estimation and path following control were integrated into the PX4 Autopilot, running on a Pixhawk 4 \ac{FMU}. The pre-computed trajectory primitives were stored on the SD card \revision[]{of the \ac{FMU}, and the air speed was measured using a pitot tube.} \revision[]{The position and attitude controllers operate at \SI{50}{\hertz} and \SI{200}{\hertz}, respectively.}

\begin{table}[t]
    \centering
    \caption{Parameter values used in the experiments.}
    \label{tab:parameters}
    \setlength{\tabcolsep}{2.0pt} 
    \renewcommand{\arraystretch}{1} 
    \begin{tabular}{c|cccccccccc}
        \toprule
        \textbf{Param} & $m$ & $A_\text{wing}$ & $C_{L0}$ & $C_{L1}$ & $C_{D0}$ & $C_{D1}$ & $C_{D2}$ & \revision[]{$C_{A0}$} & \revision[]{$C_{A1}$} & \revision[]{$C_{B1}$} \\
        \midrule
        \textbf{Value} & 1.35 & 0.4 & \revision[]{0.356} & \revision[]{2.354} & \revision[]{0.029} & \revision[]{0.378} & \revision[]{1.984} & \revision[]{0.279} & \revision[]{4.313} & \revision[]{-3.464}  \\
        \bottomrule
    \end{tabular}
    \vspace{-2ex}
\end{table}

\subsubsection{Wind Field Estimation}

The wind estimation module \revision[from Fig.~\ref{fig:architecture} (local and global estimator)]{} was evaluated in a flight test in light shear conditions. While it is difficult to obtain accurate ground-truth measurements of the wind field, we focus on the overall plausibility and consistency of the estimated wind field. For this purpose, the \ac{UAV} was manually piloted in a shear layer behind a mountain ridge, altering altitude to cover the vertical distribution. The estimates show variation in the local measurements across \revision[]{different} heights (Fig.~\ref{fig:shear_estimator_chasseral}), which results from the coarse local estimation model \eqref{wind estimation} and\revision[the]{} spatial and temporal variations in the wind field. Regardless, the global parameter estimates converge to a near-static equilibrium.

\begin{figure}[t]
\centering
    \begin{subfigure}[t]{\linewidth}
    \includegraphics[width=\textwidth]{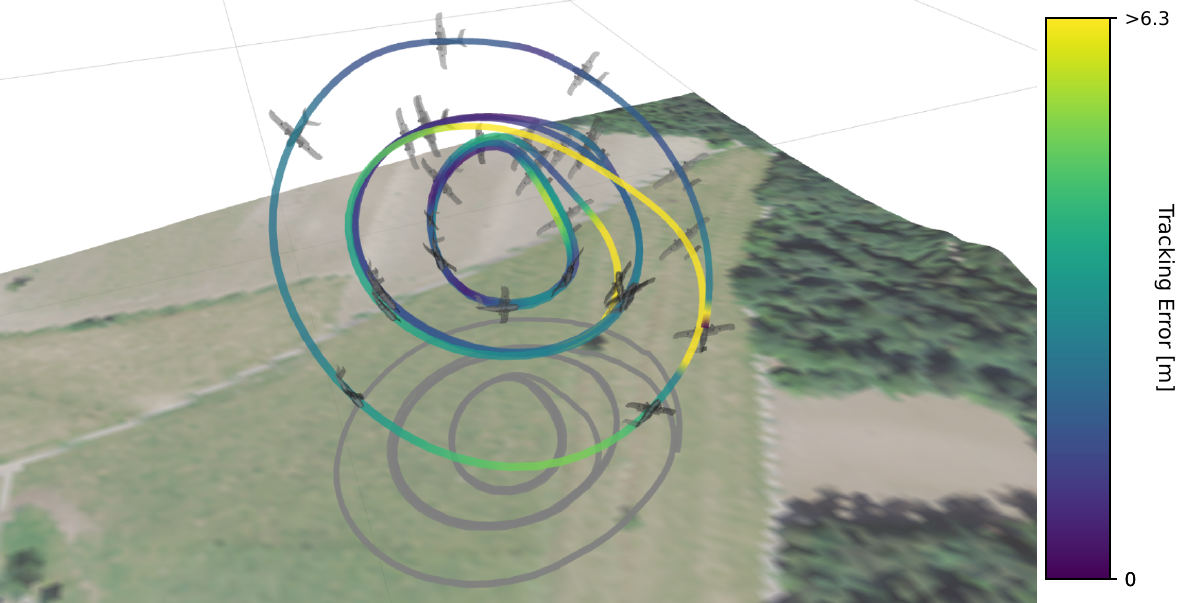}
    \subcaption{Path following arbitrary ellipse path (Z\"urich, Switzerland)}
    \end{subfigure}
    \begin{subfigure}[t]{\linewidth}
    \includegraphics[width=\textwidth]{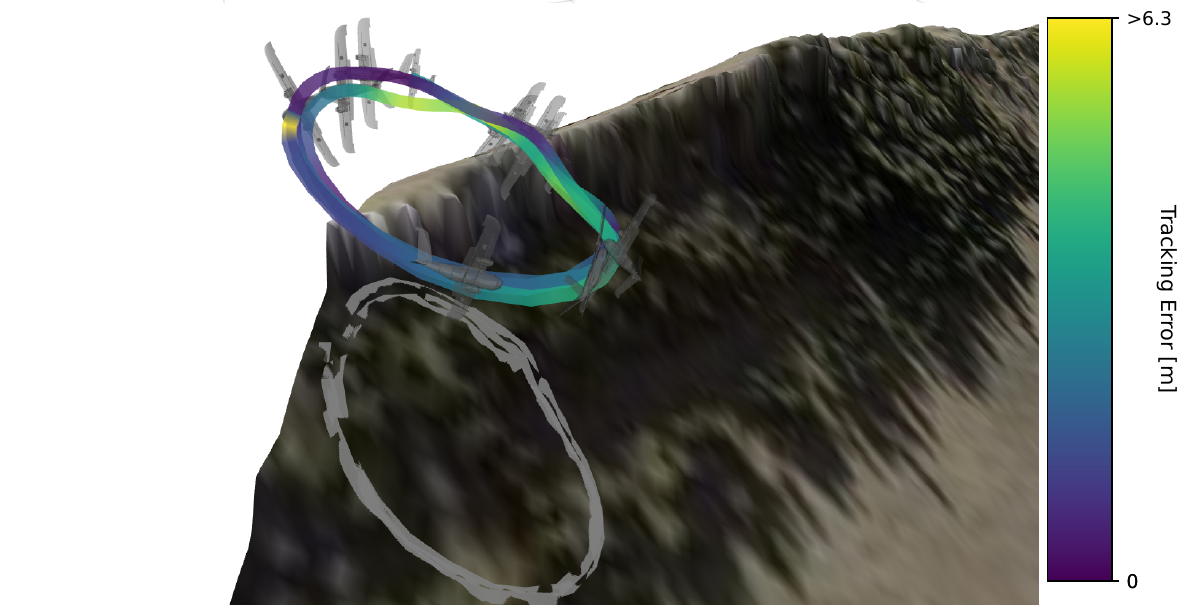}
    \subcaption{Path following strong winds (Chasseral, Switzerland)}
    \end{subfigure}
    \caption{Vehicle trajectory colored with tracking error with ground track as grey (a) arbitrary inclined ellipse paths and electric propulsion (b) strong winds. Vehicle attitude is enlarged 5 times for visibility.}\label{fig:position_tracking_path1}
\end{figure}

\subsubsection{Robust Path Following}
In the first flight experiment, the capability \revision[of the control architecture ]{}to track inclined paths with high-acceleration turns was evaluated. To imitate paths tracked in a dynamic soaring scenario, a set of inclined ellipses of different sizes and inclination \revision[angles ]{}were used as a reference path. In addition, the path of one robust soaring trajectory obtained from \eqref{robust OCP_implementation_split}\revision[ (for an arbitrary wind field and initial velocity)]{} was \revision[added to the set of paths]{used}. The test was conducted \revision[using]{by setting} the onboard electric motor \revision[set at]{to} a constant throttle to produce sufficient thrust to maintain flight in low-wind conditions ($||\textbf{w}||\leq 3\frac{m}{s}$), while switching between the reference paths\revision[. The resulting flight path and tracking error is shown in Fig.~\ref{fig:position_tracking_path1}]{~(Fig.~\ref{fig:position_tracking_path1})}. While the \revision[covered]{} state trajectories in this experiment do not represent dynamic soaring, the provided reference paths demand aggressive maneuvering, \revision[similar to]{essential for} dynamic soaring. The control architecture demonstrates good path following, with smooth switching between reference paths. The resulting tracking error largely remained below \SI{3}{\metre} \revision[]{(RMSE=\SI{2.24}{\meter})}, apart from the initial convergence phase after switching references. 

To demonstrate the tracking performance of the controller subject to strong gusting, a similar experiment was also carried out in strong wind behind a ridge (Fig.~\ref{fig:overview}, Fig.~\ref{fig:position_tracking_path1}). The present wind field offered wind speeds between \SI{8}{\metre\per\second} and \SI{12}{\metre\per\second}, with gusts up to \SI{14}{\metre\per\second}. The results highlight the ability of the control architecture to deliver solid tracking performance \revision[]{(RMSE=\SI{2.46}{\meter})} at airspeeds up to \SI{32}{\metre\per\second}. 
\revision[The performance of the tracking controller is crucial because precise maneuvering at high speed is needed for satisfactory performance in unknown and rapidly changing wind conditions (as when crossing the shear layer). 
]{}
It should be noted that the controller performance is independent of an accurate global wind estimate since only the instantaneous, local wind estimates are used for wind correction, while the global estimates are only used for planning purposes. The control performance further demonstrates small sim-to-real discrepancy with respect to path following precision. 

\subsubsection{Closed Loop Efficiency}
\begin{figure}[t]
    \centering
    \includegraphics[width=\linewidth]{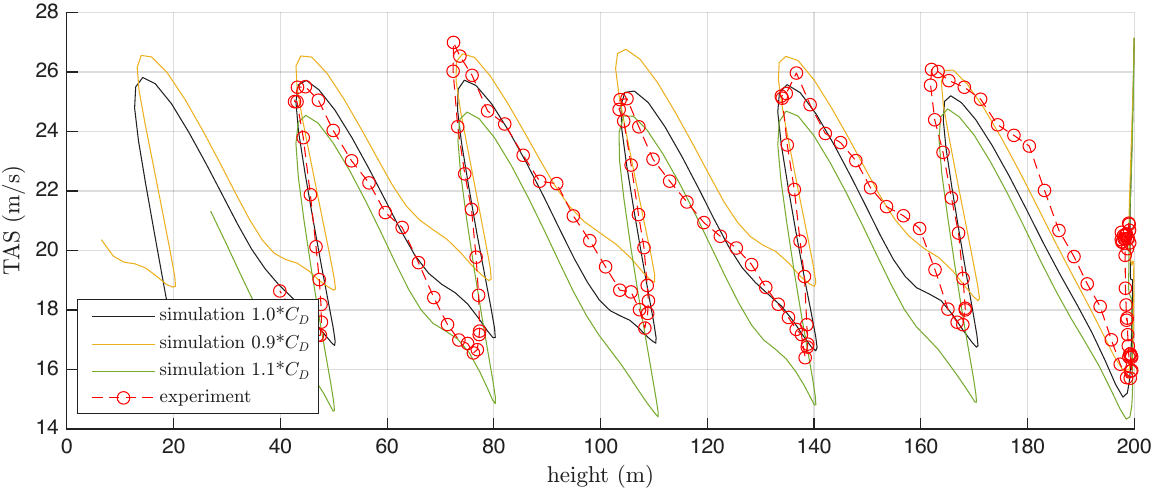}
    \caption{Airspeed evolution over height of the unpowered spiral test. The simulation results with perturbed drag coefficients are plotted in yellow and green.}
    \label{fig:spiral_airspeed_2}
\end{figure}

Finally, we \revision[need to ]{}verify that close tracking of a desired path\revision[, as demonstrated in the previous experiments,]{} also results in the expected energy evolution on the real system. This \revision[is to verify]{verifies} that the sim-to-real gap is also small with respect to potential losses from, e.g., unknown drag terms or aggressive maneuvering. 
\revision[To this end, w]{W}e compare the airspeed evolution in unpowered flight, where the system tracks an aggressive spiral-like trajectory. The flight is performed in low-wind conditions to focus on model-error induced effects, noting that robustness to wind estimation errors is shown in simulation. The airspeed evolution is compared to the simulated case in Fig.~\ref{fig:spiral_airspeed_2}. Note that the airspeed is sensitive to the aerodynamic performance during this test, as illustrated by two simulated trajectories with perturbed drag coefficients that differ significantly from the nominal case with the expected parameters. Thus, any model-error or control-induced loss of airspeed is expected to become visible during the test. Results, however, show a close match between the real flight test and the nominal simulation case. This indicates that the closed-loop performance of the pipeline closely matches the simulation.

\section{CONCLUSIONS}
In this article, we propose and demonstrate a framework for robust autonomous dynamic soaring. We show that, by using a optimization approach which considers wind uncertainty, the dynamic soaring approach can be more robust against wind uncertainty. The optimized trajectories are precomputed, and selected based on the estimated wind conditions. We demonstrate \iac{INDI}-based path following controller that can reliably track these paths.
We believe that this is a step towards demonstrating autonomous dynamic soaring on \iac{UAV}.
\revision[]{A complete \textit{in-the-loop} hardware validation of the proposed framework are left for future work due to the need for an optimized experimental platform. Further, the integration of secondary objectives like distance coverage and topographic constraints are relevant points for future work.}








\bibliographystyle{IEEEtran}
\bibliography{bibliography/references}

\end{document}

%% file: include/acronyms.tex
\DeclareAcronym{UAV}{
  short = UAV,
  long  = unmanned aerial vehicle,
  short-indefinite = a,
  long-indefinite = an
}

\DeclareAcronym{AoA}{
  short = AoA,
  long = angle of attack,
  short-indefinite = an,
  long-indefinite = an
}
\DeclareAcronym{AoS}{
  short = AoS,
  long = angle of sideslip,
  short-indefinite = an,
  long-indefinite = an,
}

\DeclareAcronym{FMU}{
  short = FMU,
  long  = flight management unit,
  short-indefinite = an,
  long-indefinite = a
}

\DeclareAcronym{GP}{
  short = GP,
  long = Gaussian process,
}
\DeclareAcronym{KF}{
  short = KF,
  long = Kalman filter, 
}
\DeclareAcronym{RL}{
  short = RL,
  long = reinforcement learning,
}
\DeclareAcronym{EKF}{
  short = EKF,
  long = extended Kalman filter,
}
\DeclareAcronym{ENU}{
  short = ENU,
  long = east north up,
}
\DeclareAcronym{FRD}{
  short = FRD,
  long = front right down,
}
\DeclareAcronym{LQR}{
  short = LQR,
  long = linear quadratic regulator,
}
\DeclareAcronym{MPC}{
  short = MPC,
  long = Model Predictive Control,
}
\DeclareAcronym{NLP}{
  short = NLP,
  long = non-linear program,
}
\DeclareAcronym{OCP}{
  short = OCP,
  long = optimal control problem,
}
\DeclareAcronym{INDI}{
  short = INDI,
  long = incremental nonlinear dynamic inversion,
  short-indefinite = an,
  long-indefinite = an
}

%% file: include/newcommands.tex
\newcommand{\reffig}[1]{Fig.~\ref{#1}}

\newcommand{\refequ}[1]{Eq.~\eqref{#1}}